\documentclass{article}
\usepackage{spconf,amsmath,graphicx}

\usepackage{times}
\usepackage{helvet}
\usepackage{courier}
\usepackage{diagbox}
\usepackage{flexisym}
\usepackage{algorithm}
\usepackage{algorithmic}
\usepackage{subfig}
\usepackage{bm}
\usepackage{amsfonts,amssymb}
\usepackage{amstext}
\usepackage{times}  
\usepackage{helvet}  
\usepackage{courier}  
\usepackage{url}  
\usepackage{graphicx}  
\usepackage{amsthm}

\usepackage{url}            
\usepackage{booktabs}       
\usepackage{amsfonts}       
\usepackage{nicefrac}       
\usepackage{microtype}      

\usepackage{tabularx}
\usepackage{amsmath,amssymb,amscd,epsfig,amsfonts,rotating}
\usepackage{graphicx}
\usepackage{epstopdf}
\usepackage{multirow}
\usepackage{bbold}
\usepackage{booktabs}
\usepackage{bbm}
\usepackage{xcolor}
\usepackage{algorithm}
\usepackage{algorithmic}
\usepackage{url}
\usepackage{amsmath,amscd,amsbsy,amssymb,latexsym,url,bm}
\usepackage{epsfig,epstopdf}
\usepackage{tipa}
\usepackage{tabu}
\usepackage{subfig}
\usepackage{amsfonts}
\usepackage{amssymb}
\usepackage{amsmath}
\usepackage{bbm}
\usepackage{bm}
\usepackage{diagbox}
\usepackage{bigints}
\usepackage{mathrsfs}
\usepackage{enumitem,balance}
\usepackage{wrapfig}
\usepackage{mathrsfs, euscript}
\usepackage{algorithm}
\usepackage{algorithmic}
\usepackage{amssymb}
\usepackage{ifpdf}
\usepackage{makecell}
\usepackage{caption}

\newenvironment{shrinkeq}[1]
{ \bgroup
  \addtolength\abovedisplayshortskip{#1}
  \addtolength\abovedisplayskip{#1}
  \addtolength\belowdisplayshortskip{#1}
  \addtolength\belowdisplayskip{#1}}
{\egroup\ignorespacesafterend}

\newtheorem{theorem}{Theorem}

\title{Adversarial Variational Bayes Methods for\\ Tweedie Compound Poisson Mixed Models}
\name{Yaodong Yang\sthanks{Correspondence to: $<$\texttt{yaodong.yang@aig.com}$>$.},  Rui Luo, Yuanyuan Liu}
\address{American International Group Inc.}
\begin{document}
%
\vspace{-5pt}

\maketitle
\vspace{-5pt}

\begin{abstract}
 The Tweedie Compound Poisson-Gamma model is routinely used for modeling non-negative continuous data with a discrete probability mass at zero. Mixed models with random effects account for the covariance structure related to the grouping hierarchy in the data. An important application of Tweedie mixed models is pricing the insurance policies, e.g. car insurance. However, the intractable likelihood function, the unknown variance function, and the hierarchical structure of mixed effects have presented considerable challenges for drawing inferences on Tweedie. In this study, we tackle the Bayesian Tweedie mixed-effects models via variational inference approaches. In particular, we empower the posterior approximation by implicit models trained in an adversarial setting. To reduce the variance of gradients, we reparameterize  random effects, and integrate out one local latent variable of Tweedie. We also employ a flexible hyper prior to ensure the richness of the approximation.
 Our method is evaluated on both simulated and real-world data. Results show that the proposed method has smaller estimation bias on the random effects compared to traditional inference methods including MCMC; it also achieves a state-of-the-art predictive performance, meanwhile offering a richer estimation of the variance function. 
 \end{abstract}

\vspace{-5pt}
\section{Introduction}
\label{sec:intro}
\vspace{-5pt}

\begin{figure}[t!]
\vskip -10.pt
\begin{center}
\subfloat[Tweedie Density Function]{\includegraphics[width=0.49\linewidth]{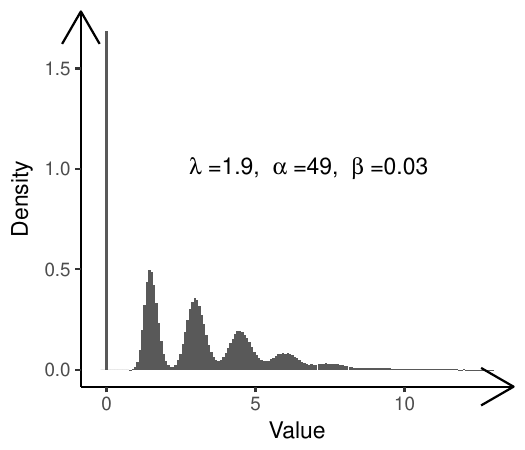} \label{td}}
\subfloat[Tweedie Mixed-Effects model.]{\includegraphics[width=0.49\linewidth]{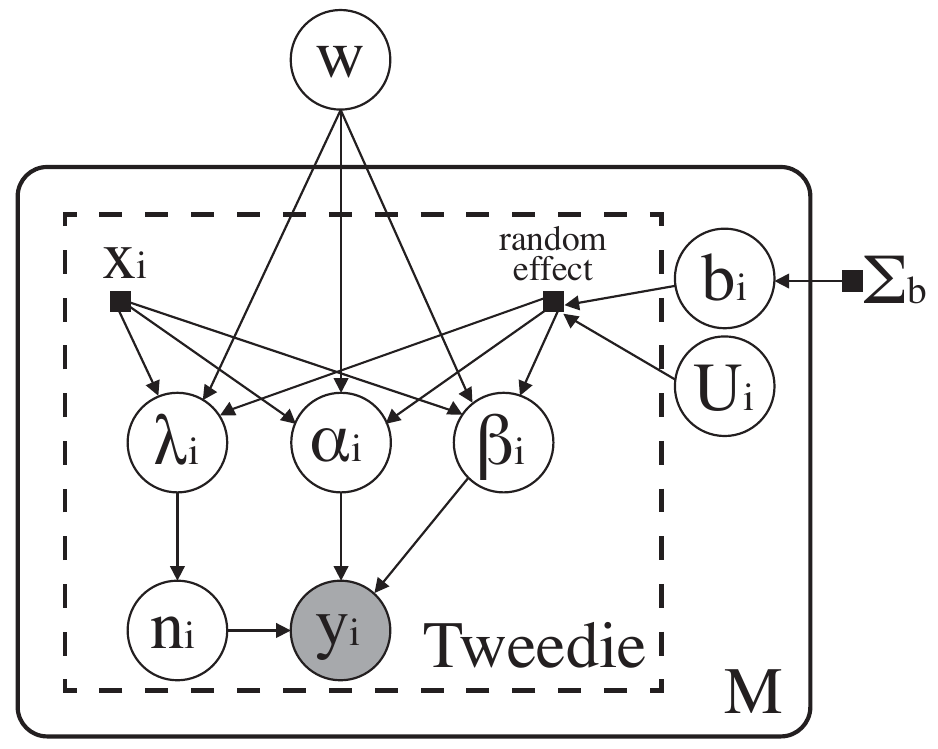} \label{gm}}
\vskip -5pt
\caption{Graphical model of Tweedie mixed-effect model.}
\label{tcpgm}
\end{center}
\vskip -25.pt
\end{figure}

Tweedie models \cite{jorgensen1997theory,jorgensen1987exponential} are special members in the exponential dispersion family;
they specify a power-law relationship  between the variance and the mean: $\text{Var}(Y) = \text{E}(Y)^{\mathcal{P}}$.
For arbitrary positive $\mathcal{P}$, the index parameter of the variance function, outside the interval of $ (0,1)$, Tweedie corresponds to a particular stable distribution, \emph{e.g.}, Gaussian ($\mathcal{P}=0$), Poisson ($\mathcal{P}=1$),  
Gamma ($\mathcal{P}=2$), Inverse Gaussian ($\mathcal{P}=3$). 
When $\mathcal{P}$ lies in the range of $(1,2)$, the Tweedie model is equivalent to Compound Poisson--Gamma Distribution  
\cite{simsekli2013learning}, hereafter \emph{Tweedie} for simplicity. Tweedie serves as a special Gamma mixture model, with the number of mixtures determined by a Poisson-distributed random variable, parameterized by$\{\lambda, \alpha, \beta\}$ and denoted as:
$Y = \sum_{i=1}^{T}G_i, T\sim\text{Poisson}(\lambda), G_i \overset{i.i.d}{\sim}\text{Gamma}(\alpha, \beta). $
Tweedie is heavily used for modeling non-negative heavy-tailed continuous data with a discrete probability mass at zero (see Fig. \ref{td}). 
As a result,
Tweedie gains its importance from multiple domains \cite{jorgensen1991exponential,smyth2002fitting}, 
including actuarial science (aggregated loss/premium modeling),  ecology (species biomass modeling), meteorology (precipitation modeling). On
 the other hand, in many field studies that require manual data collection, for example in insurance underwriting, the sampling heterogeneity from a hierarchy of groups/populations has to be considered. Mixed-effects models can represent the covariance structure related to the grouping hierarchy in the data by assigning common random effects to the observations that have the same level of a grouping variable; therefore, estimating the random effects is also an important component in Tweedie modeling. 
 
 Despite the importance of Tweedie mixed-effects models, 
the intractable likelihood function, the unknown variance index $\mathcal{P}$, and the hierarchical structure of mixed-effects (see Fig.~\ref{gm})   hinder the inferences on Tweedie models.
Unsurprisingly, 
there has been little work  devoted to full-likelihood based inference on the Tweedie mixed model, 
let alone Bayesian treatments.
In this work, 
we employ variational inference to solve Bayesian Tweedie mixed models.
The goal is to introduce an accurate and efficient inference algorithm to estimate the posterior distribution of the fixed-effect parameters, 
the variance index, and the  random effects. 


%

\vspace{-5pt}
\section{Related Work}
\label{sec:related-work}
\vspace{-5pt}

To date, most practice of Tweedie modeling are conducted within the quasi-likelihood (QL) framework \cite{wedderburn1974quasi} where the density function is approximated by the first and second order moments.  QL does not come with the schema of estimating the variance function, even though the estimation  of  the index parameter plays an important role in hypothesis testing and uncertainty evaluation \cite{davidian1987variance}. Extended quasi-likelihood (EQL) \cite{nelder1987extended}, together with the profile likelihood method \cite{cox1993note} is proposed to fill the gap of index estimation for QL. Nonetheless, EQL is not capable of handling exact zero values. 

It was not until recently that numerical approximations to the Tweedie density function appeared. Well-known approaches include series expansion method \cite{dunn2005series} and Fourier inversion method \cite{dunn2008evaluation}. Yang et al. \cite{yang2015boosted} and Chen et al. \cite{chen2016xgboost} applied the gradient boosted trees that try to incorporate non-linear interactions in estimating the mean value of the Tweedie model. Qian et al. \cite{qian2016tweedie} investigated the grouped elastic net and Lasso methods for the Tweedie model. Nonetheless, the dependence on density approximations also makes the above algorithms vulnerable to numerical issues. Both series expansion and Fourier inversion methods are noted to suffer from  unbounded increment in the number of required terms. 

When the Tweedie model is expended to incorporate the random effects, the profiled quasi-likelihood (PQL) method still suffers from the numerical restrictions. Likelihood-based inference methods require  approximation of the integral over the distribution of the random effect. Advances in numerical approximation like Laplace approximation \cite{tierney1986accurate} and adaptive Gauss-Hermite quadrature (AGQ) \cite{liu1994note} are then applied to solve the mixed model. However the complex density of the Tweedie model  stills bothers the profiled likelihood methods.

The expectation of a latent variable for an intractable likelihood function can typically be estimated using the EM algorithm in an empirical Bayes approach or MCMC methods in full Bayesian approach. Simsekli et al. \cite{simsekli2013learning,zhang2013likelihood} proposed solving the Tweedie mixed-effect models by explicitly exploiting the latent variable in the Bayesian formulation. Both MCMC and Monte Carlo EM methods have been applied. Zhang et al. \cite{zhang2013likelihood} compared the latent variable approach with the density function approximations aforementioned, and found that MCMC are computationally demanding on the high dimensional dataset, and dedicated supervisions are needed in order to adjust the Monte Carlo error and the sampling variance 
\cite{gamerman2006markov}.

\vspace{-5pt}
\section{Tweedie Mixed-Effects Models}
\vspace{-5pt}


Tweedie EDM $f_{Y}(y | \mu, \phi, \mathcal{P})$ equivalently describes the compound Poisson--Gamma distribution  when $\mathcal{P}\in(1,2)$. 
Tweedie assumes Poisson arrivals of the events,
and Gamma--distributed ``cost" for each individual event.
Judging on whether the aggregated number of arrivals  $N$ is zero, 
the joint density for each observation can be written as:
\begin{shrinkeq}{-5pt}
\begin{align}
P(Y, N=n| \lambda, \alpha, \beta)
=  & {d}_{0}{(y)} \cdot e^{-\lambda} \cdot  \mathbbm{1}_{n=0}  \nonumber \\ + &  \frac{y^{n\alpha -1}e^{-y/\beta}}{\beta^{n\alpha}\Gamma(n\alpha)} \cdot  \frac{\lambda^n e^{-\lambda}}{n!}  \cdot \mathbbm{1}_{n>0}\label{jcpgd},
\end{align}
\end{shrinkeq}
where  $d_{0}(\cdot)$ is the Dirac Delta function at zero.
The connection of the parameters $\{\lambda, \alpha, \beta\}$ of Tweedie Compound Poisson model with the parameters of the general Tweedie  EDM model, parameterized by the mean, the dispersion and the index parameter $\{\mu, \mathcal{P}, \phi \}$ is denoted as:
\begin{align}
\left\{\begin{matrix} 
\lambda=&\frac{\mu^{2-\mathcal{P}}}{\phi(2-\mathcal{P})}\\[.5em]
\alpha=&\frac{2-\mathcal{P}}{\mathcal{P}-1}\\[.5em]
\beta=&\phi(\mathcal{P}-1)\mu^{\mathcal{P}-1}
\end{matrix}\right.\quad,\quad
\left\{\begin{matrix}
\mu =& \lambda \alpha \beta\\[.5em]
\mathcal{P}=&\frac{\alpha+2}{\alpha+1}\\[.5em]
\phi=&\frac{\lambda^{1-\mathcal{P}}(\alpha\beta)^{2-\mathcal{P}}}{2-\mathcal{P}}.
\end{matrix}\right. 
\label{mapping}
\end{align}
A mixed-effects model contains both fixed effects and random effects; graphical model is shown in Fig.~\ref{gm}.
We denote the mixed model as:
$\kappa(\mu_i)=f_{\bm{w}} (\bm{X}_{i})+\bm{U}_{i}^{\top}\bm{b}_{i}, 
\bm{b}_{i}{\sim}\mathcal{N}(\bm{0, \Sigma_b})$
where $\bm{X}_{i}$ is the $i$-th row of the design matrix of  fixed-effects covariates, $\bm{w}$
are parameters of the fixed-effects model which could represent linear function or  deep neural network.
$\bm{U}_{i}$ is the $i$-th row of the design matrix associated with random effects, 
$\bm{b}_{i}$ is the coefficients of the random effect which is usually  assumed to follow a 
multivariate normal distribution with zero mean and covariance
$\bm{\Sigma_b}$, $\kappa(\cdot)$ is the link function,
and $\mu_{i}$ is the mean of the $i$-th response variable $Y$.
 In this work, 
we have considered the random effects on the intercept.



In the context of conducting Bayesian inference on Tweedie mixed models, 
we define 1) the observed data $D=(\bm{x}_{i}, \bm{u}_{i}, y_{i})_{i=1, \ldots, M}$.
2) global latent variables $\{\bm{w},\bm{\sigma}_{\bm{b}}\}$, we assume 
$\bm{\Sigma_b}$ is a diagonal matrix with its diagonal elements
$\bm{\sigma}_{\bm{b}}$; 3) local latent variables $\{n_i, \bm{b}_{i}\}$, 
indicating the number of  arrivals and the random effect. 
The parameters of Tweedie is thus denoted by $(\lambda_i, \alpha_i, \beta_i)=f_{\lambda, \alpha, \beta}(\bm{x}_{i}, \bm{u}_i| \bm{w}, \bm{\sigma}_{\bm{b}})$.
The latent variables thus contain both local and global ones $\bm{z}=(\bm{w}, n_i, \bm{b}_{i}, \bm{\sigma}_{\bm{b}})_{i=1,\ldots,M}$, 
and they are assigned with prior distribution $P(\bm{z})$. 
The joint log-likelihood is computed by summing over the number of observations $M$ by $ \sum_{i=1}^M \log \left[ P(y_i | n_i, \lambda_i, \alpha_i, \beta_i)\cdot P(n_i | \lambda_i)\cdot P(\bm{b}_i|\bm{\sigma}_{\bm{b}}) \right]$.
The goal here is to find the posterior distribution of $P(\bm{z}|D)$ and  make future predictions via $P(y_{\text{pred}}|D, \bm{x}_{\text{pred}}, \bm{u}_{\text{pred}} )=\int P(y_{\text{pred}}|\bm{z}, \bm{x}_{\text{pred}}, \bm{u}_{\text{pred}})  P(\bm{z}|D) \operatorname {d}\bm{z}$.

\vspace{-5pt}

\section{Methods}
\vspace{-5pt}


\textbf{Adversarial Variational Bayes.}
Variational Inference (VI) \cite{jordan1999introduction} 
approximates the posterior distribution, 
often complicated and intractable, 
by proposing a class of probability distributions $Q_\theta({\bm{z}})$ (so-called inference models), 
and then finding the best set of parameters $\theta$ by minimizing the KL divergence between the proposal and the true distribution, \emph{i.e.}, $\text{KL}(Q_\theta({\bm{z}})||P(\bm{z}|D))$. 
Minimizing the KL divergence 
is equivalent to maximizing the evidence of lower bound (ELBO) \cite{bishop2006pattern},  
 expressed as Eq.~\ref{bound}.
Optimizing the ELBO requires the gradient information of $\nabla_{\theta}\text{ELBO}$.  In our experiments, 
we find that the model-free gradient estimation method -- REINFORCE  \cite{williams1992simple} 
 fails to yield reasonable results due to the unacceptably-high variance issue \cite{hinton1995wake},
 even equipped with \emph{baseline} trick 
\cite{mnih2014neural} or \emph{local expectation} \cite{aueb2015local}. We also attribute the unsatisfactory results to the over-simplified proposal distribution in traditional VI. 
Here we try to solve these two issues by employing the implicit inference models with variance reduction tricks.
\begin{shrinkeq}{-8pt}
\begin{equation}
\theta^* = \arg\max_\theta \operatorname {E}_{Q_{\theta}(\bm{z})} \bigg[ -\log \frac{Q_\theta(\bm{z})}{P(\bm{z})} + \log P(D|\bm{z}) \bigg]\label{bound}	
\end{equation}
\end{shrinkeq}
AVB 
\cite{huszar2017variational,mescheder2017adversarial} empowers the VI methods by using neural networks as the inference models; 
this allows more efficient and accurate approximations to the posterior distribution.  
Since a neural network is black-box,
the inference model is implicit thus have no closed form expression, even though this does
not bother to draw samples from it. 
To circumvent the issue of computing the gradient from implicit models, 
the mechanism of adversarial learning is introduced; 
an additional discriminative network $T_{\phi}(\bm{z})$ is used to model the
 first term in Eq.~\ref{bound}. By building a model 
to distinguish the latent variables that are sampled from the prior distribution $p(\bm{z})$ 
from those that are sampled from the inference network $Q_{\theta}(\bm{z})$, 
namely, 
optimizing the blow equation (where $\sigma(x)$ is the sigmoid function):
\begin{shrinkeq}{-6pt}
\begin{align}
\phi^*=\arg\min_\phi \bigg[- & \operatorname {E}_{Q_\theta(\bm{z})} \log \sigma(T_\phi(\bm{z})) \nonumber \\ - &  \operatorname {E}_{P(\bm{z})} \log (1 - \sigma(T_\phi(\bm{z})) \bigg],
\label{discriminator}
\end{align}
\end{shrinkeq}
 the ratio is estimated as $\operatorname E_{Q_\theta(\bm{z})}[  \log \frac{Q_\theta(\bm{z})}{P(\bm{z})} ] =\operatorname  E_{Q_\theta(\bm{z})} [ T_{\phi^*}(\bm{z}) ]$.
AVB considers optimizing Eq.~\ref{bound} and Eq.~\ref{discriminator} as a two-layer minimax game. 
We apply stochastic gradient descent alternately to find a Nash-equilibrium. 
Such Nash-equilibrium, 
if reached, 
is a global optimum of the ELBO 
\cite{mescheder2017adversarial}.

\textbf{Reparameterizable Random Effect.}
In mixed-effects models, the random effect is conventionally assumed to be  $\bm{b}_{i}{\sim}\mathcal{N}(\bm{0,\Sigma_b})$. In fact, they are reparameterisable. 
	As such, we model the random effects by the reparameterization trick
\cite{kingma2013auto}; $\bm{b}_{i}$ is now written as $\bm{b}(\bm{\epsilon}_{i})= \bm{0} + \bm{\sigma}_{\bm{b}} \odot \bm{\epsilon}_{i}$ where $\bm{\epsilon}_{i}\sim\mathcal{N}(\bm{0,\bm{I}})$. 
The  $\bm{\sigma}_{\bm{b}}$
is a set of latent variables generated by the inference network and the random effects become \emph{deterministic} given the auxiliary noise $\bm{\epsilon}_{i}$.
\begin{theorem}[Reparameterizable Random Effect]
Random effect is naturally reparameterizable. With the reparameterization trick, the random effect is no longer restricted to be normally distributed. For any ``location-scale" family distributions (Students't, Laplace, Elliptical, \emph{etc.})
\begin{proof}
See the Section 2.4 in \cite{kingma2013auto}
\end{proof}
\end{theorem}
\vskip -5pt
 Note that when computing the gradients, we no longer need to sample the random effect directly, instead, we can back-propagate the path-wise gradients  which could dramatically reduce the variance \cite{kucukelbir2016automatic}.


\newcommand\T{\rule{0pt}{2.6ex}}       
\newcommand\B{\rule[-1.2ex]{0pt}{0pt}}

\textbf{Hyper Priors.}
The priors of the latent variables are fixed in traditional VI.
Here
we however parameterize the prior $P_\psi (\bm{w})$ by $\psi$ and make $\psi$ also trainable when optimizing Eq.~\ref{bound}. 
We refer $\psi$ as a kind of \emph{hyper prior} to $\bm{w}$. 
The intuition is to not constraint the posterior approximation by one over-simplified prior.
We would like the prior to be adjusted so as to make the posterior $Q_\theta(\bm{z})$ close to a set of prior distributions,
or a self-adjustable prior; this could further ensure the expressiveness of $Q_\theta(\bm{z})$. We can also apply the same trick if the class of prior is reparameterizable.




\textbf{Variance Reduction.}
We find that integrating out the latent variable $n_i$ by Monte Carlo in Eq.~\ref{jcpgd} gives significantly lower variance in computing the gradients.
As $n_i$ is a Poisson generated sample, 
the variance will explode in the cases where $Y_{i}$ is zero but the sampled $n_i$ is positive, 
and $Y_i$ is positive but the sampled $n_i$ is zero. 
This also accounts for why the direct application of REINFORCE algorithm fails to work. In practice,  we find 
limiting the number of Monte Carlo samples
between $2 - 10$,  dependent on the dataset,  has the similar performance as summing over to larger number.
\begin{shrinkeq}{-6pt}
\begin{align}
&P(y_{i}|\mathbf{w},\mathbf{b},\mathbf{\Sigma})\nonumber\\
&= P(\mathbf{b},\mathbf{\Sigma})\cdot\sum_{j=1}^T P(y_i|n_j, \mathbf{w},\mathbf{b},\mathbf{\Sigma}) \cdot P(n_j|\mathbf{w},\mathbf{b},\mathbf{\Sigma}) \label{reconstruction}
\end{align}
\end{shrinkeq}
\vskip -10pt


\begin{algorithm}[h] 
                \caption{AVB for Bayesian Tweedie mixed model}
                \label{alg}
                \begin{algorithmic}[1]
                                \STATE {\bfseries Input:} data $D=(\mathbf{x_i},\mathbf{u_i}, y_i)$, $i=1,..., M$
                                \WHILE {$\bm\theta$ not converged}
                                                \FOR {$N_T$:}
                                                \STATE Sample noises $\bm{\epsilon}_{\bm{i}, \bm{j} =1,\ldots,M} \sim N(\mathbf{0},\mathbf{I})$, 
                                                \STATE Map noise to prior $\mathbf{w}_{j}^P= P_{\bm\psi}(\bm{\epsilon_j}) = \bm{\mu} + \bm{\sigma} \odot \bm{\epsilon_{j}}$, \\
                                                \STATE Reparameterize random effect 
                                    $\bm{b}_{j}^P =\bm{0}+\bm{\sigma_b} \odot \bm{\epsilon_{j}}$
                                                \STATE Map noise to posterior $\mathbf{z}_{i}^Q = (\mathbf{w}_{i},\mathbf{b}_{i})^Q= Q_{\bm\theta}(\bm{\epsilon_i})$, 
                                                \STATE Minimize Eq.~\ref{discriminator}  over $\bm\phi$ via the gradients: \\
                                                $\nabla_{\bm\phi} \frac{1}{M} \sum_{i=1}^M \left[- \log \sigma(T_{\bm\phi}(\mathbf{z}_i^Q)) - \log (1 - \sigma(T_{\bm\phi}(\mathbf{z}_i^P)) \right]$
                                                \ENDFOR
                                                \STATE Sample noises $\bm{\epsilon}_{\bm{i} =1,\ldots,M} \sim N(\mathbf{0},\mathbf{I})$,
                                                \STATE Map noise to posterior $\mathbf{z}_{i}^Q =(\mathbf{w}_{i},\mathbf{b}_{i})^Q= Q_{\bm\theta}(\bm{ \epsilon_i})$, 
                                                \STATE Sample a mini-batch of $(\mathbf{x_i}, y_i)$ from $D$,
                                                \STATE Minimize Eq.~\ref{bound} over $\bm\theta, \bm\psi$ via the gradients: \\
                                                $\nabla_{\bm\theta, \bm\psi} \frac{1}{M}\sum_{i=1}^M [ -T_{\bm\phi^*}(\mathbf{z}_i^Q) + $\\ 
                                                 $\sum_{j=1}^T P(y_i|n_j, \mathbf{w},\mathbf{b},\bm{\sigma_b}; \mathbf{x_i},\mathbf{u_i})\cdot P(n_j|\mathbf{w}; \mathbf{x_i},\mathbf{u_i})\cdot P(\mathbf{b}\mid \bm{\sigma_b}) ]  $ 
                                \ENDWHILE
                \end{algorithmic} 
                \label{algo1}
\end{algorithm}
\vskip -15pt

\begin{table*}[t!]
\vskip -25pt
			\caption{The pairwise Gini index comparison with standard error based on $20$ random splits}
	\begin{center}
	\vskip -12pt
		\begin{normalsize}
			\begin{sc}
				\resizebox{.99\textwidth}{!}{%
				\begin{tabular}{l|rrrrrrr}
					\hline
					\diagbox[width=3.8cm]{\small{Baseline}}{\small{Model}} &GLM\cite{mccullagh1984generalized}& PQL \cite{mccullagh1984generalized} &Laplace  \cite{tierney1986accurate}&AGQ\cite{liu1994note}&MCMC\cite{zhang2013likelihood}&TDBoost\cite{yang2017insurance}&AVB\\
					\hline
					GLM  (autoclaim)& $/$ & $-2.97_{6.28}$ & $1.75_{5.68}$& $1.75_{5.68}$& $-15.02_{7.06}$&$1.61_{6.32}$&$\bm{9.84_{5.80}}$\T \\
					PQL & $7.37_{5.67}$ & $/$ & $7.50_{6.26}$ & $7.50_{5.72}$&$6.73_{5.95}$&$0.81_{6.22}$&$\bm{9.02_{6.07}}$ \\ 
					Laplace &$2.10_{4.52}$ &$-1.00_{5.94}$ & $/$ & $8.84_{ 5.36}$&$4.00_{4.61}$&$21.45_{4.84}$&$\bm{20.61_{4.54}}$ \\
					AGQ& $2.10_{4.52}$  &$-1.00_{5.94}$ & $8.84_{ 5.36}$ & $/$&$4.00_{4.61}$&$21.45_{4.84}$&$\bm{20.61_{4.54}}$ \\
					MCMC &$14.75_{6.80}$&$-1.06_{6.41}$ & $3.12_{5.99}$ & $3.12_{5.99}$&$/$&$7.82_{5.83}$&$\bm{11.88_{5.50}}$ \\
TDBoost&$17.52_{4.80}$&$17.08_{5.36}$&$19.30_{5.19}$&$19.30_{5.19}$&$11.61_{4.58}$&$/$&$\bm{20.30_{4.97}}$\\			
					AVB& $-0.17_{4.70}$ & $0.049_{5.62}$ & $3.41_{4.94}$ & $3.41_{4.94}$&$0.86_{4.62}$&$11.49_{4.93}$&$/$\B \\
				    \hline
				    GLM  (fineroot)& $/$ & $23.18_{9.30}$ & $35.87_{6.52}$& $35.87_{6.52}$& $-15.73_{10.63}$&$35.71_{6.52}$&$\bm{35.64_{6.53}}$\T\\
					PQL & $ -21.61_{8.34}$ & $/$ & $ 30.90_{8.86}$ & $ 30.90_{8.86}$&$-21.45_{8.38}$&$24.19_{8.36}$&$\bm{28.97_{8.92}}$ \\ 
					Laplace &$-12.55_{7.37}$ &$-14.72_{8.81}$ & $/$ & $15.07_{ 6.41}$&$-12.55_{7.37}$&$15.33_{7.36}$&$\bm{10.61_{7.20}}$ \\
					AGQ& $-12.55_{7.37}$ &$-14.72_{8.81}$ & $15.07_{ 6.41}$& $/$&$-12.55_{7.37}$&$15.33_{7.36}$&$\bm{10.61_{7.20}}$ \\
					MCMC &$17.27_{10.25}$&$22.53_{9.31}$ & $35.10_{6.54}$ & $35.10_{6.54}$&$/$&$35.10_{6.54}$&$\bm{34.87_{6.55}}$ \\
TDBoost&$22.47_{6.80}$&$8.50_{9.09}$&$22.63_{6.80}$&$22.63_{6.80}$&$11.61_{6.80}$&$/$&$\bm{22.39_{6.80}}$\\			
					AVB& $-8.261_{7.66}$ & $-10.886_{8.98}$ & $2.13_{7.28}$ & $2.13_{7.28}$&$-8.26_{7.66}$&$11.00_{7.74}$&$/$\B \\
					\hline
				\end{tabular}}
			\end{sc}
		\end{normalsize}
	\end{center}
		\label{ginicomp}
	\vskip -15.pt
\end{table*}


\section{Experiments and Results}


We compare our method with six traditional inference methods on Tweedie. 
We evaluate our method on two public datasets of modeling the aggregate loss for auto insurance polices \cite{yip2005modeling}, and  modeling the length density of fine roots in ecology \cite{de1999analysis}. 
 We separate the dataset in $50\%/25\%/25\%$ for train/valid/test respectively. 
 Considering the sparsity and right-skewness of the data, we use the  ordered Lorenz curve and its corresponding Gini index \cite{frees2011summarizing,frees2014insurance} as the metric. Assuming for the $i_{th}/N$  observations, $y_i$ is the ground truth, $p_i$ to be the results from the baseline predictive model, $\hat{y}_i$ to be the predictive results from the model. We sort all the samples by the relativity $r_i = \hat{y_i}/{p_i}$ in an increasing order, and then compute the empirical cumulative distribution as
$(\hat{F}_p(r)=\frac{\sum_{i=1}^{n}p_i\cdot \mathbbm{1}(r_i\leqslant r)}{\sum_{i=1}^{n}p_i}, 
\hat{F}_y(r)=\frac{\sum_{i=1}^{n}y_i\cdot \mathbbm{1}(r_i\leqslant r)}{\sum_{i=1}^{n}y_i})$.
The plot of $(\hat{F}_p(r), \hat{F}_y(r))$ is an ordered Lorenz curve, and Gini index is the twice the area between the Lorenz curve and the $45^{\circ}$ line. The Gini index is not symmetrical if different baselines are used. In short, a model with a larger Gini index indicates its superior capability of separating the observations. In the insurance context, Gini index profiles the model's capability of distinguishing policy holders with different risk levels.

\begin{figure}[t!]
\vskip -0pt
\begin{center}
				 \includegraphics[width=.8\columnwidth]{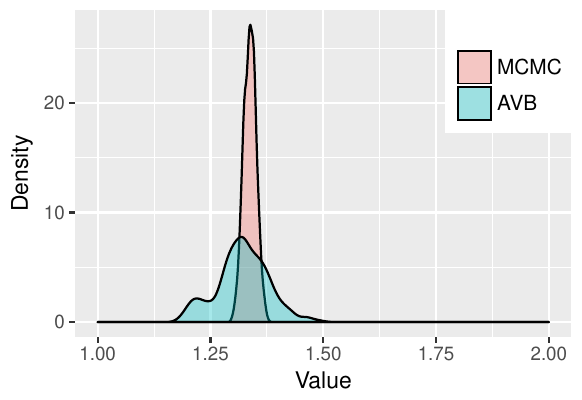} \vskip -5pt
    \captionof{figure}{Posterior estimation of the index parameter $\mathcal{P}$ on the AutoClaim data.}\label{pp}
    \end{center}
            \vskip -20pt
    \end{figure}

 Gini comparison  is presented in Table.~\ref{ginicomp}. Here only fixed effects are considered. AVB method shows the state-of-the-art performance on both AutoClaim and FineRoot datasets, even comparatively better than the boosting tree method that is considered to have a strong performance.  By inspecting the posterior estimation of $\mathcal{P}$ in Fig.~\ref{pp}, we find AVB shows a richer representation of the index parameter, with two modals at $1.20$ and $1.31$. As  $\mathcal{P}$ uniquely characterizes a Tweedie distribution, compared with one single value that traditional profiled likelihood method offers, flexible  choices of $\mathcal{P}$ enable  customized descriptions for insurance policies that may have different underlying risk profiles.   Also note that AVB uses a neural sampler  that does not involve any rejection procedures; unlike MCMC, it holds the potentials for large--scale predictions on high dimensional dataset.


   \begin{table}
   \center
   \scalebox{0.8}{
\begin{tabular}{l|rrrr}
					\toprule
					\diagbox[width=2.5cm]{\small{Algo.}}{\small{Param.}} 
										&$\mathcal{P}_{\text{auto}}$& $\sigma_{\text{b,auto}}^2$ &$\mathcal{P}_{\text{root}}$ &$\sigma_{\text{b,root}}^2$\\
					\hline
					PQL& $/$ & $8.94\cdot 10^{-5}$ & $/$& $1.0346\cdot 10^{-3}$\T \\
					LAPLACE& $1.3394$ & $4.06\cdot 10^{-4}$ & $1.4202$& $6.401\cdot 10^{-3}$\T \\
					AGQ & $1.3394$ & $4.027\cdot 10^{-4}$ & $1.4202$ & $6.401\cdot 10^{-3}$\T\\ 
					MCMC &$1.3458$ &$4.575\cdot 10^{-3}$ & $1.4272$ & $3.490\cdot 10^{-2}$\T\\
					AVB& $1.3403$  &$3.344\cdot 10^{-3}$ & $1.4237$ & $2.120\cdot 10^{-2}$\T\B \\
					\bottomrule
				\end{tabular}}
      \captionof{table}{Estimation on $\mathcal{P}$ and $\sigma_b^2$ for both AutoClaim and FineRoot data with random effects.}\label{inference_P}
      \vskip -10pt
				\end{table}

To compare the estimations on random effects,
we add ``CARTYPE" in the AutoClaim data as the random effect with 6 groups and ``PLANT" in the FineRoot data with 8 groups. 
From table \ref{inference_P}, we can see that all the algorithms estimate $\mathcal{P}$ in a reasonable similar range while the results from MCMC and AVB are closer. 
The estimation of $\sigma_b^2$ has a large variance. 
The estimation from AGQ and Laplace methods are around ten times smaller than MCMC and AVB results while PQL produces even smaller estimations. 
This is consistent with our finding in the simulation study  that AGQ and Laplace methods tends to underestimate the random effect variance. The AVB and MCMC estimations are considered more accurate; as the estimated variance is comparatively smaller than MCMC, the AVB estimation can be believed to be more accurate.

\vspace{-5pt}
\section{Conclusions}
\vspace{-5pt}
We present Adversarial Variational Bayes in solving Bayesian Tweedie mixed models. To empower the posterior approximation, we employ implicit models and training in an adversarial setting. To reduce variance of the gradients, we sum over the local latent variable of Tweedie model that is not parameterizable. We also make the prior distribution trainable to guarantee the expressiveness of the posterior. 
Our method outperforms all traditional methods on both simulated data and real-world data; meanwhile, the estimations on the variance parameter and the random effects are competitively accurate and more stable.
Tweedie model is one fundamental model that is widely used in  insurance policy pricing.
As far as we are concerned, this is the first work that leverages recent progresses of deep learning methods to tackle insurance problems.  


\newpage
\small
\bibliographystyle{IEEEbib}
\bibliography{tweedie_bib}

\end{document}